\documentclass[rn,11pt]{ucl_document} 
\usepackage{graphicx}
\usepackage{hyperref} 
\usepackage{fancyheadings} 

\newlength{\halfwidth}
\newlength{\halfwidthextra}
\newlength{\temp}
\newlength{\tempa}

\begin{document}

\title{
Long-Term Evolution of Genetic Programming Populations
}

\author{
{W. B. Langdon}
}

\date{24 March 2017}
\documentnumber{17/05}

\maketitle

\pagenumbering{arabic}  
\pagestyle{plain}

\begin{abstract} 
We evolve 
binary mux-6 trees for up to 100\,000 generations
evolving some programs with more than a hundred million nodes.
Our unbounded 
Long-Term Evolution Experiment LTEE 
GP 
appears not to evolve building blocks
but does
suggests a limit to bloat.
We do see periods of tens even hundreds of generations 
where the population is 100\% functionally converged.
The distribution of tree sizes is not 
as predicted
by theory. 

Short version appears as \cite{Langdon:2017:GECCO}.


\end{abstract}

\section{Introduction} 

\noindent
Rich Lenski's experiments in long term evolution
\cite{Lenski:2015:PRoySocB}
in which the BEACON team evolved bacteria for more than 60\,000
generations
and found continued beneficial adaptive mutations,
prompts the same question in computation based evolution.
What happens if we allow artificial evolution,
specifically genetic programming~(GP) with crossover
\cite{koza:book,banzhaf:1997:book,poli08:fieldguide},
to evolve for tens of thousands, even a hundred thousand generations.

Firstly we expect bloat%
\footnote{%
  GP's tendency to evolve non parsimonious solutions 
  has been known since the beginning of genetic programming.
  E.g.~it is mentioned in
  Jaws \cite[page~7]{koza:book}.
  \label{p.intronsbloat}
  Walter Tackett \cite[page~45]{Tackett:1994:thesis}
  credits Andrew Singleton with
  the theory that GP bloats due to the 
  cumulative increase in 
  non-functional code, known as introns.
  The theory says these
  protect other parts of the same tree by deflecting genetic
  operations from the functional code by simply offering more
  locations for genetic operations.
  The bigger the introns, the more chance they will be hit by crossover
  and so the less chance crossover will disrupt the useful part of
  the tree.
  Hence bigger trees tend to have children with higher fitness than
  smaller trees.
  See also
  \cite{kinnear:altenberg,kinnear:angeline}.},
and so we need a GP system not only able to run for
$10^{5}$~generations
but also able to process trees with in excess of a 100 million nodes.
Section~\ref{sec:results} describes our target system
which is able to do this on a Linux desktop computer.
Section~\ref{sec:size}
shows we do indeed see bloat but after a few hundred generations we
start to see surprises.
Tree growth is not continuous.
Indeed we see many generations where the trees get smaller.
As we look at the very long term distribution of fitness
in Section~\ref{sec:fit}
we do not get the whole population
reaching one fitness value
and everyone having that fitness in all subsequent generations.
Section~\ref{sec:fit}
also applies existing theory
to give a mathematical model
of fitness convergence's impact on selection.

Section~\ref{sec:introns}
discusses introns and constants.
By introns we mean
subtrees which cannot impact on the program's output,
even if modified by crossover.
A constant
is simply a 
subtree whose output is the same for all input test cases.
Although introns are well known in GP,
we find constants are also required to explain the observed long
term evolution.
In
Section~\ref{p.smallmums}
we confirm the asymmetry of the GP crossover
and the importance of the first parent,
which gives the offspring's root node
and show this becomes even more important in long term evolution.
(Since this parent typically gives more genetic material
we call it the mother, or simply mum.)
Section~\ref{sec:entropy}
looks at the distribution of values within the large evolved trees
and particularly
how combining values within the evolving trees leads to constants.

Section~\ref{sec:multifit}
considers the evolution of multiple solutions as
an alternative to classic bloat as protection
from disruptive crossover.
However it
concludes that although multiple solutions within the highly evolved
programs are possible, they do not 
explain the evolution seen.
Whereas
Section~\ref{sec:constants}
shows the fraction of constants in highly evolved trees 
is fairly stable despite huge fluctuations in the size of the trees
containing them.

Section~\ref{sec:shape}
shows the classic fractal Flajolet random distribution of trees~\cite{langdon:2000:fairxo}
still applies to highly evolved trees.
It also shows,
as expected,
the Flajolet distribution 
also applies to the
sub trees within them.
Finally Section~\ref{sec:code}
demonstrates the common assumption that useful code
clusters around the root node
and shows that in the longer term this code becomes highly stable
and is surrounded and is protected by many thousands of useless
instructions
composed of both introns and constants.

Section~\ref{sec:endbloat}
proposes a lose limit to bloat
in extremely long GP runs.
The limit turns out to be somewhat ragged.
Nevertheless Section~\ref{sec:endbloat}
provides some experimental evidence for it 
in runs lasting one hundred
thousand generations.

\subsection{Background}

\noindent
There has been some work in evolutionary computing 
and artificial life looking at the extended evolutionary process itself
rather than focusing on solving a problem.
For example,
Harvey's Species Adaptation Genetic Algorithms (SAGA)~\cite{harvey:1997:ob}
argued in favour of variable length representation
(such as we use in genetic programming)
to allow continued evolutionary adaptation.
Whilst more recently Lewis' TMBL system \cite{Lewis:2010:cec}
turned to the power of parallel processing GPUs to
investigate the long term
gathering of small mutations over many generations.

McPhee~\cite[sect.~1.2]{mcphee:2001:EuroGP} 
reported that his earlier studies which had reported the first 100
generations of the INC-IGNORE problem could not safely be extrapolated 
to long term evolution
but considered runs of up to 3000 generations.
McPhee~\cite{mcphee:2001:EuroGP} 
found ``increasing amount of noise''
in the evolution of their linear unary trees
and say it was unclear if their tree lengths would be bounded.
As in our previous work~\cite{langdon:2000:quad}
we will use binary tree GP\@.
However previously~\cite{langdon:2000:quad}
we
only considered the first few hundred generations,
whilst in the next sections
we shall investigate many tens of thousands.

\section{Results} 
\label{sec:results}

\noindent
Following \cite{langdon:2000:quad},
the generational GP was run with a population of 500
(see Table~\ref{gp.details}).
All parents were selected using independent tournaments of size seven
to select both parents used by each crossover.
One child was created by each crossover.
For simplicity, all crossover points are chosen uniformly at random.
I.e.\ there was no bias in favour of functions 
\cite{koza:book}.
Mutation was not used.
Since we will evolve trees with more than a hundred million nodes
it is essential to use a compact representation and fast fitness
evaluation,
therefore we use Poli's Sub-machine code GP
\cite{poli:1999:aigp3}
on a 64 bit processor
which allows us to process all 64 fitness cases simultaneously.

\begin{table}
\setlength{\temp}{\textwidth}
\settowidth{\tempa}{Fitness cases:}
\addtolength{\temp}{-\tempa}
\addtolength{\temp}{-2\tabcolsep}
\caption{\label{gp.details}
Long term evolution of 6-Mux trees
}
\begin{tabular}{@{}lp{\temp}@{}}\hline
Terminal set: \rule[1ex]{0pt}{6pt} & D0 D1 D2 D3 D4 D5 \\
Function set:                      & AND OR NAND NOR \\
Fitness cases:                     & All $2^6$ combinations of inputs D0..D5\\
Selection:  & tournament size 7 using the number of correct fitness cases
\\ 
Population: & 500. Panmictic, non-elitist, generational.
\\ 
Parameters: &
Initial population created at random using ramped half and half
\cite{koza:book} with depth between 2 and 6.
100\%~subtree crossover~\cite{koza:book}.
10\,000~generations
(runs cut short on any tree reaching hard limit of a million nodes).
\\\hline
\end{tabular}
\end{table}

\subsection{Evolution of Size}
\label{sec:size}

\begin{figure} 
\centerline{\includegraphics[scale=1.175]{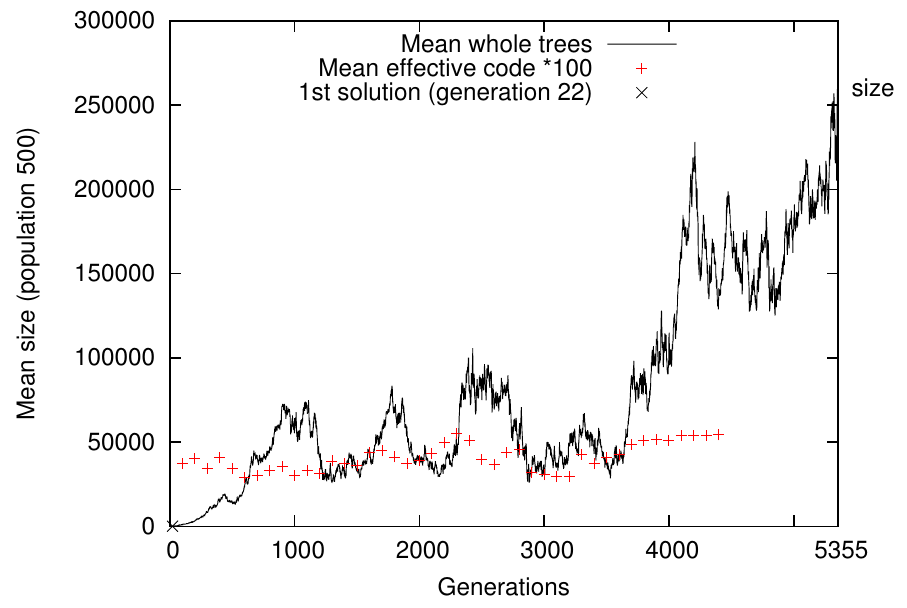}} 
\vspace*{-2ex}
\caption{\label{fig:bmux6_100_3L2_size}
Evolution of tree size in first 6-mux run.
(This run 
aborted after 5355 gens by first crossover to hit
million node limit.)
Because of computational cost, the 
average number of effective instructions
(i.e.~not introns or blocked by constants)
is plotted~($+$) every 100 gens.
(Mean 
effective code
497.)
}
\end{figure}

\noindent
In all cases we do see enormous increases in size
(see Figure~\ref{fig:bmux6_100_3L2_size}).
Except for runs in Section~\ref{sec:endbloat},
in each case the run was cut short because bloat became so severe that
further crossovers where inhibited by the hard size limit
of a million nodes (see Table~\ref{gp.details}).
In the initial generations,
we do see explosive growth in tree size
and this continues even after a tree with max fitness is found.
Indeed bloat continues even as the first time 
everyone in the population has the same fitness value
($1^{\rm st}$ convergence).

If everyone in the population has the same fitness,
selection appears to become random
and children are as likely to be smaller than their parents
as they are likely to be bigger.
If selection is switched off for many generations,
we do indeed see an apparently random walk in average tree size,
with falls and rises.
However tree size cannot go negative
and once the population contains only small trees
crossover cannot escape and the trees remain tiny forever.
That is, we have a gambler's ruin.
(We return to this in Section~\ref{sec:endbloat}.)

\begin{figure}
\centerline{\includegraphics[scale=1.175]{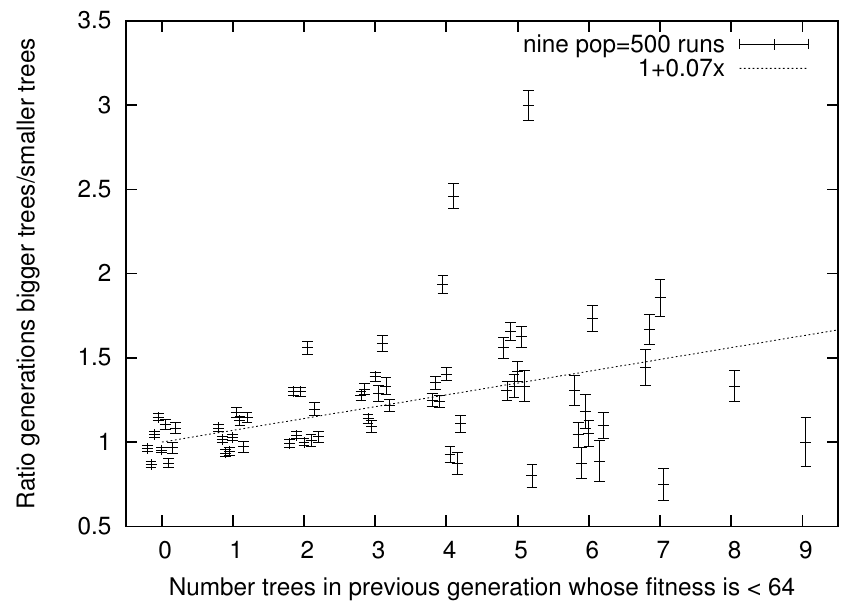}} 
\vspace*{-2ex}
\caption{\label{fig:updown}
Impact of poor fitness individuals on average tree size in next generation.
Ratio of cases where mean tree size increased to where it fell
v.~number of trees with fitness $<64$ in parent generation.
Nine runs after first time 100\% population has max fitness.
Dotted line is median linear regression.
}
\end{figure}

Although after the time where almost everyone in the population
has the same fitness value,
we do see falls in average tree size as well as increases,
this is not a simple random walk.
For example,
in Figure~\ref{fig:bmux6_100_3L2_size} there are substantial 
falls in size which take place over many generations
during which
the number of generations in which the trees are smaller on average
v.\ the number in which they are bigger, is too large to be simply random.
This might be due to the discovery of 
smaller than average individuals of max fitness
but with a higher than average effective fitness,
however we were unable to find hard evidence to support this hypothesis.
(Effective fitness is simply fitness rescaled to take into account the
disruptive effects of crossover and mutation
\cite[sec.~14.2]{nordin:thesis},
\cite[page~187]{banzhaf:1997:book}
\cite{Stephens:1999:ECJ},
\cite{langdon:fogp}.)
Even after the population has converged to the point
where everyone has the same fitness
($1^{\rm st}$ convergence),
crossover can still be disruptive
so there are generations with lower fitness individuals.
Since they never have children but 
(as we will see in Section~\ref{p.smallmums} 
on page~\pageref{p.smallmums})
they tend to be smaller than average,
they lead to increases in tree size.
Figure~\ref{fig:updown}
shows the ratio of generations where average tree size increased
to where it decreased
v.\
the number of unfit trees in the parent
generation.
To limit noise,
we do not plot
data with $\le 5$ instances
where size increased or
where it decreased.

After the population $1^{\rm st}$ converges, tournament selection has no problem
immediately removing 
unfit children and so
in later generations
there are either none or very few of them.
Thus the average tree size is largely the same as in the previous
generation plus some random variation.
Nonetheless over thousands of generations the removal of smaller trees
is sufficient to continue to bloat the population.

\subsection{Fitness Convergence}
\label{sec:fit}

\begin{figure}
\centerline{\includegraphics[scale=1.175]{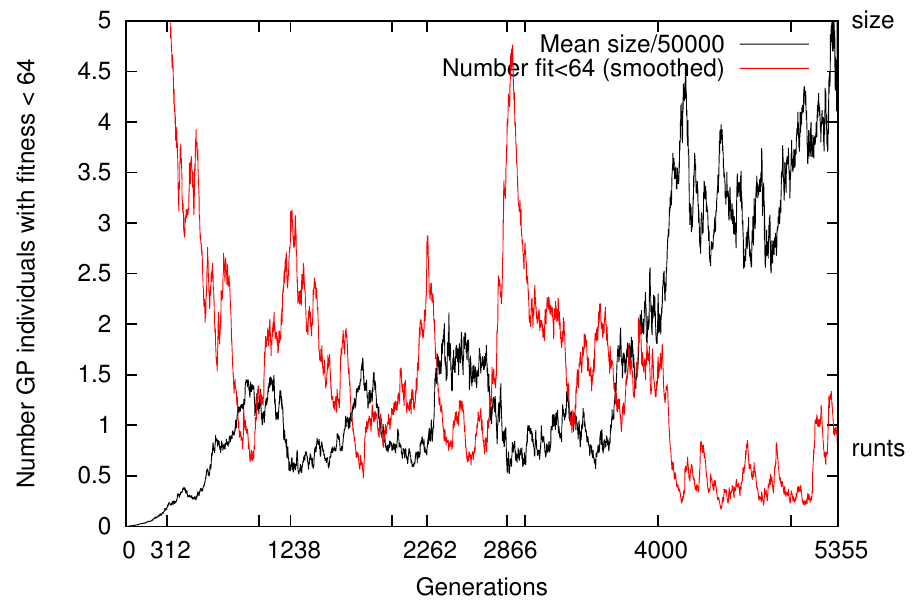}} 
\vspace*{-2ex}
\caption{\label{fig:bmux6_100_fitness}
Evolution of low fitness in first 6-mux pop=500 run.
All poor fitness trees (runts)
first removed in gen~312.
(Plot has been smooth over 
30 generations.) 
After generation 312, there are at most 11 trees with fitness $<64$
(gens 1238, 2866
and one 10,  gen 2262).
(Mean prog size in background.)
}
\end{figure}

\noindent
At the start of the run better trees evolve
which
tournament selection chooses to be
the parents for many crossovers.
These often succeed in propagating the parents'
abilities to their children.
So, as expected, the number of individuals with maximum fitness
in the population grows rapidly
towards 100\%.
However as
Figure~\ref{fig:bmux6_100_fitness} shows,
typically it does not remain at 100\%
but hovers slightly below 100\%.
There is a good deal of stochastic fluctuation and so 
Figure~\ref{fig:bmux6_100_fitness} shows the value smoothed
over thirty generations.
A third of generations after generation~312
also only contain trees with max fitness.
But almost as many~(28\%) have one low fitness individual
and 18\% have two.
Figure~\ref{fig:bmux6_100_fitness} also shows a general 
downward trend towards fitness convergence
as the trees tend to get bigger without a corresponding increase in the
size of their effective code 
(shown with $+$ in Figure~\ref{fig:bmux6_100_3L2_size}).

\begin{figure} 
\centerline{\includegraphics[scale=1.175]{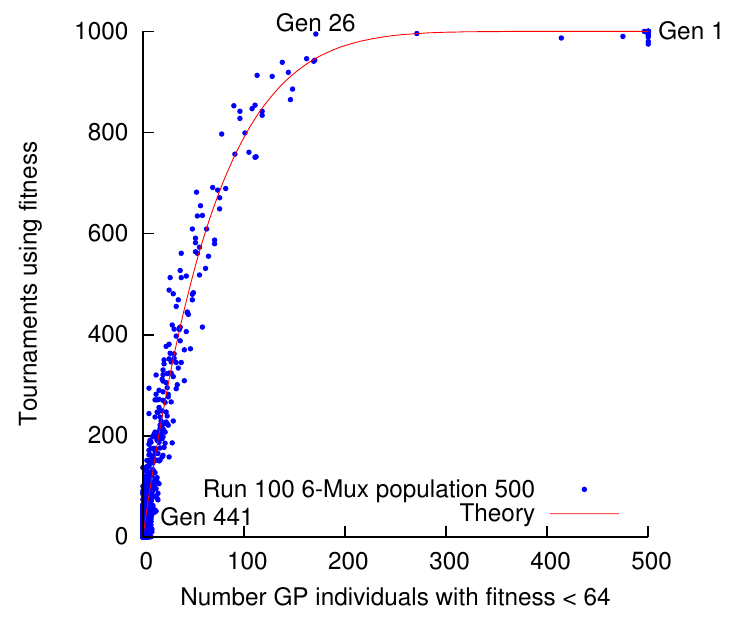}} 
\vspace*{-2ex}
\caption{\label{fig:bmux6_noselection}
Convergence of typical GP population leads to reduction in fitness
variation and so fewer tournaments using any aspect of fitness.
Except where a population converges to fitness below 64,
all runs
follow the theoretical curve 
\mbox{$2\times {\rm popsize} \left(1-(1-x/{\rm popsize})^{\rm tourn\ size}\right)$}.
}
\end{figure}

Figure~\ref{fig:bmux6_noselection},
(page~\pageref{fig:bmux6_noselection})
shows 
once a GP run has found a tree with max fitness,
the impact of fitness selection on subsequent generations falls rapidly
in line with theory.
I.e.\ the expected number of tournaments $y$ using some aspect of fitness
is given by
\mbox{$y=2\times {\rm popsize} \left(1-(1-x/{\rm popsize})^7\right)$}
where $x$ is the number of poor fitness trees in the population
and the tournament size is 7.
Near the origin 
$y \approx 14 x$.
In a typical run, after first fitness convergence,
two thirds of generations contain one, two or more poor children.
Meaning on average in $2/3^{\rm rds}$ of populations,
fitness selection plays some role in the choice of
parents for 
14, 28 or more children.

\subsection{Introns}
\label{sec:introns}

\noindent
The functions available to evolution are AND, OR,
NAND and NOR\@.
If, for all fitness test cases, an input to an AND node is~0, 
then the AND's output will also always be~0.
Whereas if an input to
NAND is always~0
NAND's output is always~1.
Similarly always~1 for OR gives always~1, 
and it gives always~0 for NOR\@.
Our GP system does not pre-define such constants,
however they can be readily evolved in many ways.
E.g.\
0 can be created by taking any leaf and anding it with its inverse.
Given these four combinations of functions and 0 and 1,
the other input to the function has no impact on the nodes output.
Indeed, since our function set has no side-effects,
the whole subtree leading to
the input has no impact.
Some GP systems may be optimised to avoid even evaluating it.
Further, not only can it have no effect in the current tree,
any genetic changes to it also have no effect.
Meaning children who only differ from their parent in 
such subtrees are guaranteed to have exactly the same fitness as their
parent.
In genetic programming we have been calling such subtrees ``introns''.
Although it does not use this information,
our GP system was modified to recognise and report such introns.

\subsection{The Importance of Mothers}
\label{p.smallmums}

\begin{figure*}
\begin{tabular}{@{}c@{\hspace{\columnsep}}c@{}}
\rule{\halfwidth}{0pt} &
\rule{\halfwidth}{0pt}\\
\includegraphics[scale=1.18607]{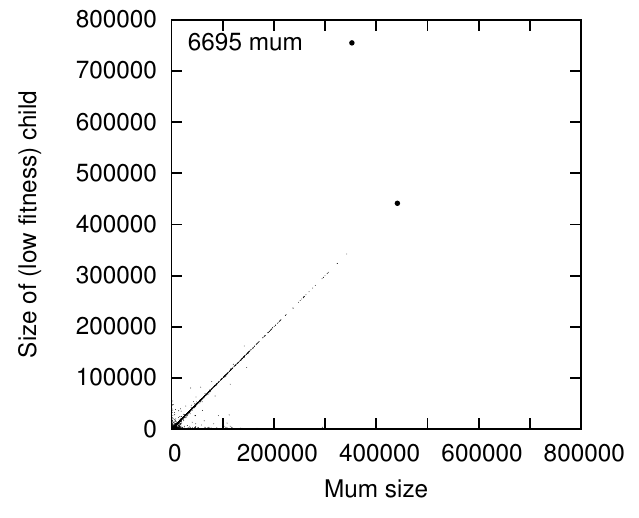} 
&
\includegraphics[scale=1.18607]{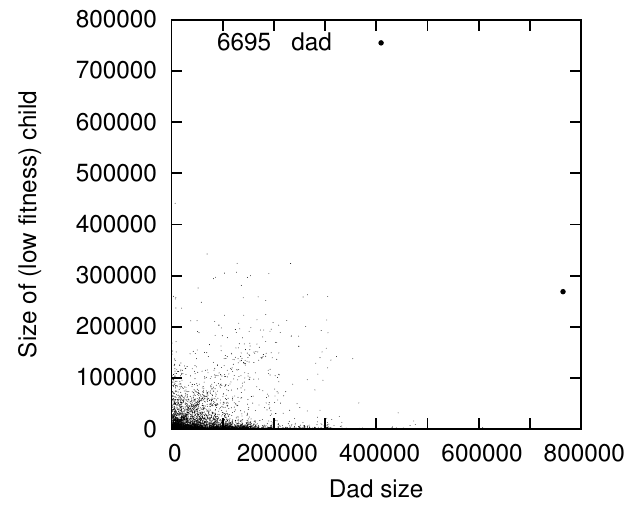} 
\end{tabular}
\vspace*{-2ex}
\caption{\label{fig:bmux6_100_runt}
Size of all 6695 children (typical run) 
with fitness less than their parents
compared to size of their parents (x-axis)
which both have max fitness~(64).
Note strong correlation with root node parent 
(mum).
}
\end{figure*}

\begin{figure*}
\begin{tabular}{@{}c@{\hspace{\columnsep}}c@{}}
\rule{\halfwidth}{0pt} &
\rule{\halfwidth}{0pt}\\
\includegraphics[scale=1.2]{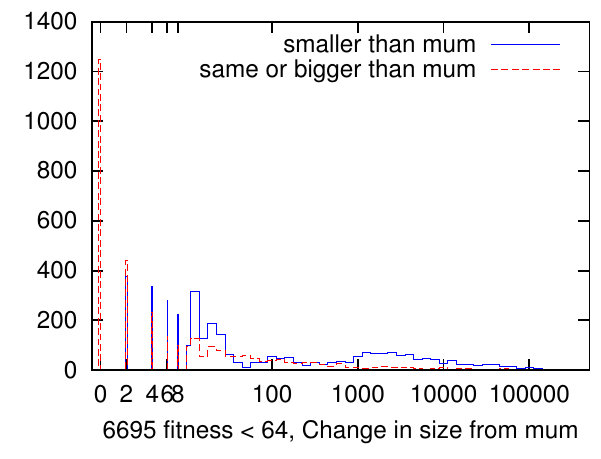} 
&
\includegraphics[scale=1.2]{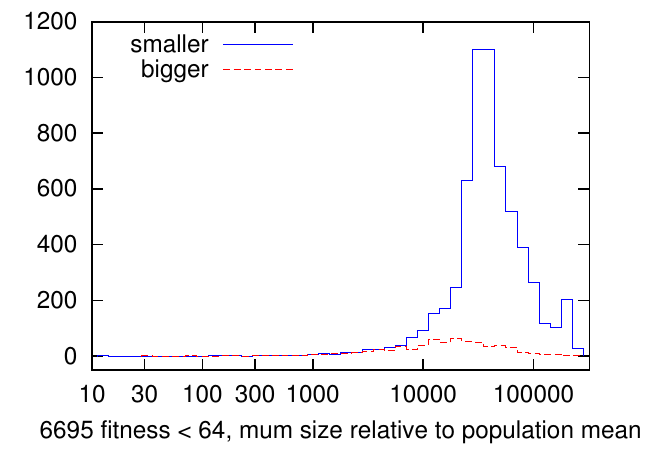} 
\end{tabular}
\vspace*{-2ex}
\caption{\label{fig:bmux6_100_runt_hist}
Left:
Change in size of 6695 poor fitness children (typical run) 
(Decile bins used for changes bigger than 8).
More than half
(3428) of the children are smaller than their 1st parent,
2018~(30\%) larger 
and 1249~(19\%) the same size.
\label{fig:bmux6_100_6_mum}
Right:
Distribution of size of 1st parent (mum) relative to population mean
for all 6695 low fitness ($<64$) children (typical run) 
where both parents have max fitness~(64).
90\% 
of the mums 
are smaller than average.
(Same data as Figure~\protect\ref{fig:bmux6_100_runt}.)
}
\end{figure*}

\noindent
Figure~\ref{fig:bmux6_100_runt}
shows the size of
low fitness children is highly correlated with the parent
(mother) which they inherent their root node from.
Figure~\ref{fig:bmux6_100_runt_hist}~(left) shows that 
crossovers which change fitness
tend to remove more code than they add
but this effect is dwarfed by the factor that 
90\% of mothers of unsuccessful
offspring are smaller than average
and on average the difference is much bigger 
than the change in size caused by
the damaging crossover,
see Figure~\ref{fig:bmux6_100_6_mum} (right).
However 
(as we shall see in the next section page~\pageref{p.bmux6_100_xo})
in many cases the fraction of the new population with worse
fitness than their mothers
is far smaller than the fraction of introns
and the difference is largely made up by the protection afforded by
constants.

\subsection{Entropy \& Evolution of Robust Constants}
\label{sec:entropy}

\noindent
Although all the components in
the evolved trees are deterministic
they lead to progressive information loss 
(entropy reduction).
When the components are deeply nested
(i.e.~the trees are large)
the loss may be complete,
leading to
subtrees 
whose output is independent
of their leafs.
We call such zero entropy subtrees constants.
$$\mbox{Entropy is }\ \
S= -64\left(p \log_2(p) + q \log_2(q)\right)$$
where $p$ is the fraction of the 64 fitness cases which are true
(and $q$ for zero).
We use $\log_2$ so that entropy is expressed in bits.
Notice in these experiments
the entropy of all leafs is 64 bits 
and all constants have entropy of zero.
With our,
deliberately symmetric,
function set,
the entropy of a function with identical inputs
is the same as that of those inputs.
Whilst the entropy of a function whose inputs are two different leafs
will be
$S= -64\left(\frac{3}{4}\log_2\left(\frac{3}{4}\right) + \frac{1}{4}\log_2\left(\frac{1}{4}\right)\right) = 51.92
\ \mbox{bits} $
I.e.\ lower than that of either input.
This is generally true of any program:
the entropy of each step is typically lower than the entropy of its
inputs and cannot be higher.
(This is reasonable since no deterministic system can
increase the information content of its input.)
Another way to view this is that the impact of each leaf is diluted
in each function by the effect of the leafs attached
(possible indirectly)
to the function's other input.
This suggests leafs in a function's subtree but separated from it by
many intermediate levels of the subtree would tend to have less impact
on it.
If the path from any code to the root node passes through a zero
entropy function,
it cannot effect the output of the tree 
and we call it ineffective code.

Large evolved trees
contain evolved constants. 
For example both trees in
Figure~\ref{fig:bmux6_100_id_200000}
(page~\pageref{fig:bmux6_100_id_200000})
contain three constants 
created by OR functions.
The whole of the subtree below the constant can be replaced by the
constant without changing the program's output in any way.
Evolved
constants are resilient to changes in the subtree 
beneath them.
In a typical run
after the first 
time the population converges
so that everyone has maximum fitness
(i.e.\ generation 312) 
only one or two crossovers made in subtrees headed by a
constant reduce the child's fitness.
\label{p.bmux6_100_xo}
On average over 500 generations after generation 312,
99.6\% of children are modified only in code below a constant,
on average in each generation
of these 
1.9
(0.38\% of the population)
do not have maximum fitness.
The remaining 0.4\% of crossovers, give rise to on average 0.8
poor fitness children
(0.16\% of the population).

\subsection{Evolution of High Fitness Within Trees}
\label{sec:multifit}

\begin{figure}
\centerline{\includegraphics[scale=1.175]{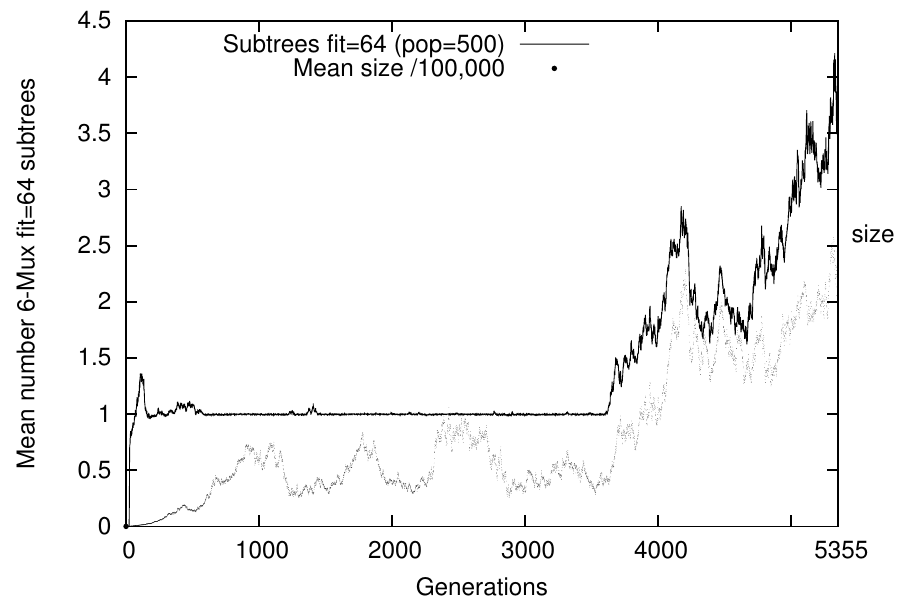}} 
\vspace*{-2ex}
\caption{\label{fig:bmux6_100_nsol}
Evolution of average number of 6-mux subtrees in the population 
with max fitness~(64)
(typical GP run).
Notice several thousand generations where there is almost exactly one
per tree in the population.
Only after extreme bloat (dotted line)
do we sometimes
get $>1$ 
subtrees with fit=64.
}
\end{figure}

\noindent
A potential alternative way for GP individuals to protect their
children in the evolving population might be to contain 
multiple instances of high fitness code
so that crossover between descendants
would have some chance of copying complete
high fitness code (building blocks).
This would also require the insertion point to be receptive.
However if we take the view that the
high fitness code must itself be a subtree,
then Figure~\ref{fig:bmux6_100_nsol} shows this does not happen.
In all our pop=500 runs,
even after thousands of generations,
on average
there are less than a handful of 
subtrees per individual which themselves have max fitness.
Since they are so few
they stand very little chance of being selected as a
crossover point in trees of several hundred thousand nodes.

\begin{figure}
\centerline{\includegraphics[scale=1.175]{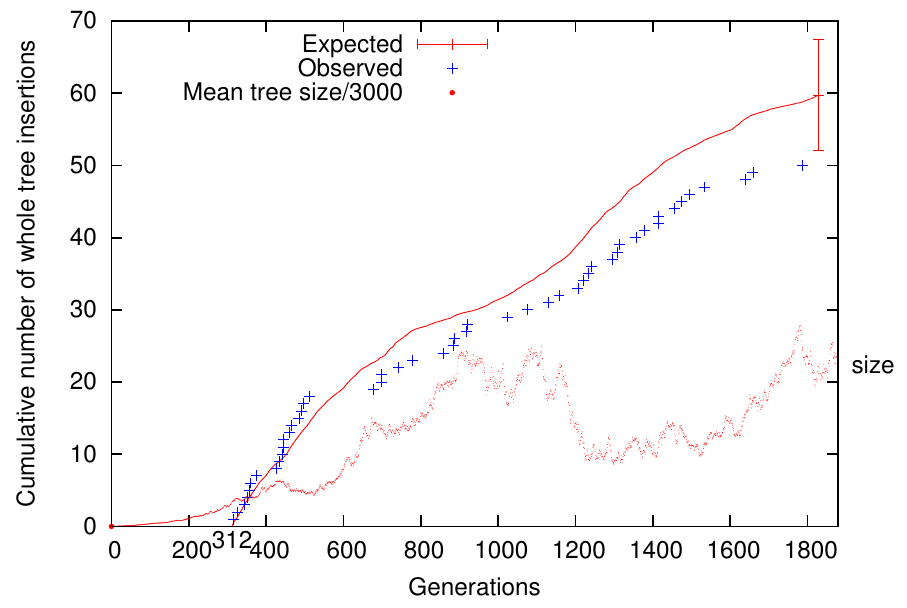}} 
\vspace*{-2ex}
\caption{\label{fig:pTraceDynamic_anal}
Count of number of whole trees inserted~$+$
(i.e.\ crossover point at root of second parent).
Showing agreement with theory (solid line).
The count is started from the first time the population 
converges to 100\% max fitness
and is shown for the next 500 generations.
For ease of comparison the average size of trees is shown dotted.
The expected rate of increase falls as trees get bigger.
}
\end{figure}

Figure~\ref{fig:pTraceDynamic_anal} 
(page~\pageref{fig:pTraceDynamic_anal})
shows good agreement between
the observed occurrence of whole tree insertion crossovers
and theory based on the size of the trees.
Between 
gen~312 (the first time the whole population reached max fitness) 
and 
generation 1828
there were
758\,500 crossovers.
Based on tree sizes, about 60 
should have caused the whole of one parent to be inserted into the other.
Although they give rise to children which contain multiple 
subtrees with fitness~64
these do not spread through the population
and the average number of such subtrees remains near~1.0
until much later
(see Figure~\ref{fig:bmux6_100_nsol}).

\subsection{Evolution of Constants}
\label{sec:constants}

\begin{figure}
\centerline{\includegraphics[scale=1.175]{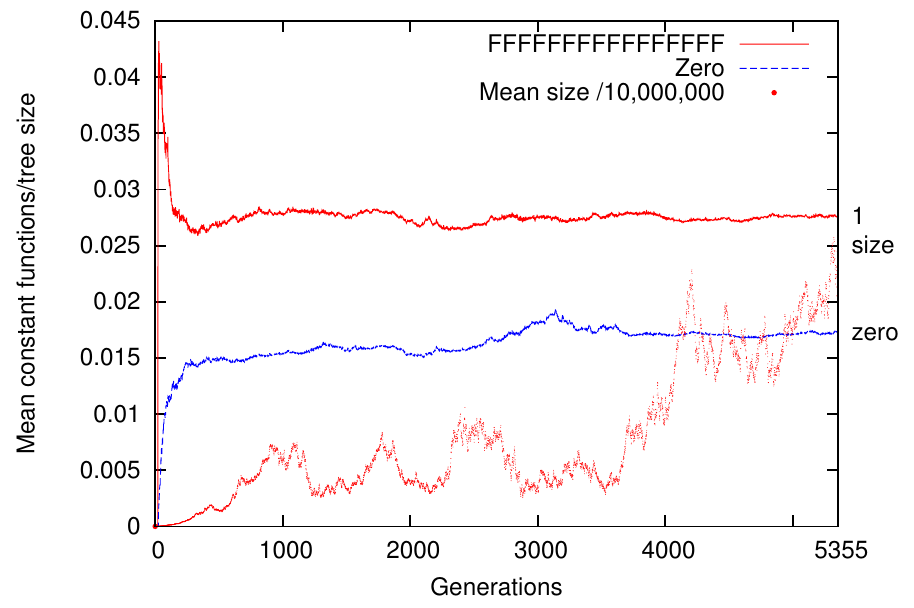}} 
\vspace*{-2ex}
\caption{\label{fig:bmux6_100_constant}
Evolution of average fraction of nodes in the population 
which evaluate to a constant
(typical GP run).
Notice several thousand generations where the
fraction of the evolving population which are constant is almost unchanging.
}
\end{figure}

\noindent
With our \cite{langdon:2000:quad}
Boolean function set,
evolution readily assembles constants
and these rapidly spread through the population.
The exact ratios vary between runs.
(Figure~\ref{fig:bmux6_100_constant},
page~\pageref{fig:bmux6_100_constant},
shows their evolution in a typical run.)
After a few hundred generations,
functions which evaluate to the same value for all test cases
occupy a few percent of the whole population
and once evolved these fractions are fairly stable to the end of the run.
Such high densities of constants reflect the large fraction of introns
and ineffective code in the highly evolved trees.

\subsection{Evolution of Tree Size and Depth}
\label{sec:shape}

\begin{figure}
\centerline{\includegraphics[scale=1.42857]{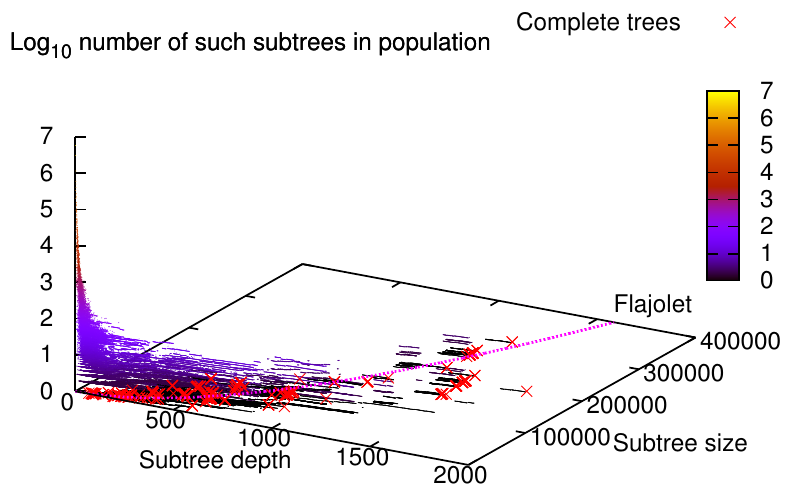}} 
\vspace*{-2ex}
\caption{\label{fig:bmux6_100_sized}
Generation 2500 of typical 6-mux run.
Highly evolved trees ($\times$) lie close to Flajolet line (dotted parabola).
As predicted their subtrees are also like random trees
\protect\cite{langdon:1999:sptfs}.
}
\end{figure}

\noindent
Figure~\ref{fig:bmux6_100_sized} 
(page~\pageref{fig:bmux6_100_sized})
shows the distribution of subtrees
within a highly evolved population at generation 2500.
As expected
\cite{langdon:1999:sptfs},
in no case are trees either maximally short and bushy or 
maximally tall and thin.
Instead both trees and subtrees lie near the 
mean size v.~depth limit
calculated by Flajolet for random binary trees of a given size~%
\cite{flajlet:1982:ahbt}.
Random trees have the fractal like property that often 
there is a leaf close to their root node and
this is also true of subtrees within them.

\begin{figure*}
\begin{tabular}{@{}p{\halfwidth}p{\halfwidth}@{}}
\includegraphics{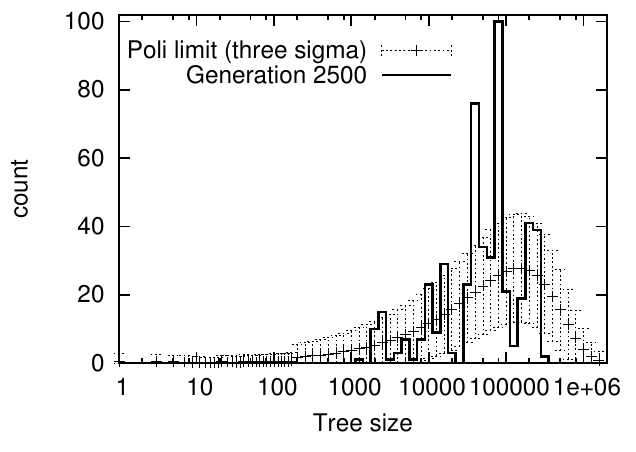} 
&
\includegraphics{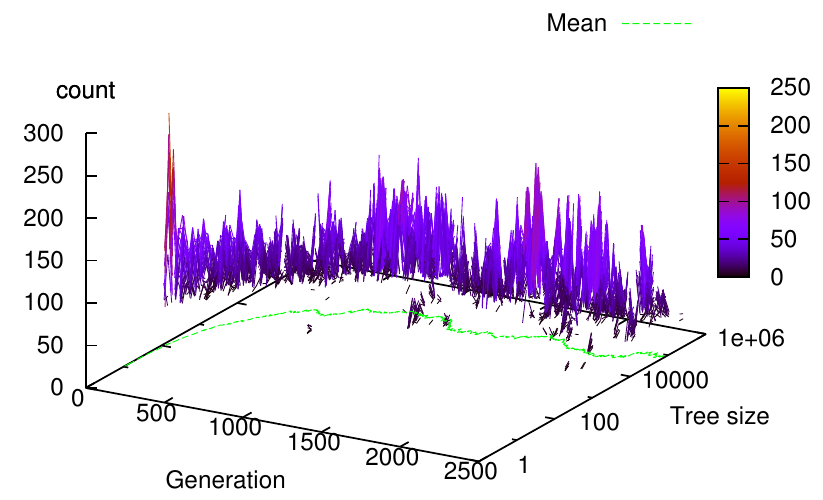} 
\end{tabular}
\vspace*{-2ex}
\caption{\label{fig:bmux6_100_poli}
Distribution of highly evolved fit trees in a typical run
(decile bins).
Left:
Generation 2500.
The dotted line shows
Poli's theoretical limiting distribution for random crossover
\cite[Eq~24]{poli:2007:eurogp}
(arity $a=2$, $p_{a}$ set to fit observed mean: 82482.1).
Right:
Generations 0 to 2500.
Colour indicates by how much each decile bin exceeds 
value plus
three standard deviations
predicted by
Poli's limiting distribution \cite{poli:2007:eurogp}.
Note log scales.
}
\end{figure*}

If we compare the evolved distribution of tree sizes
with Poli's limiting distribution
\cite{poli:2007:eurogp}
the match is good but
the actual distribution does differ significantly from theory,
see Figure~\ref{fig:bmux6_100_poli}.
Nonetheless the theory does predict both the extended tail to very
small trees and the upper tail.
It also predicts reasonably well the location of the peak.

The generation (2500) depicted in Figure~\ref{fig:bmux6_100_poli}~(left)
is not atypical.
If we look at all generations leading up to it,
after generation 35,
in every case
at some point one or more groups of trees have more similar sizes than 
predicted by unfettered random crossover
(colour in Figure~\ref{fig:bmux6_100_poli} right).
Suggesting even the very modest degree of fitness selection continues
to have an impact on the population (even when the trees get smaller
with time).
Again the distribution predicts the tails reasonable well.

\subsection{Evolution of Effective Code}
\label{sec:code}

\begin{figure*}
\setlength{\temp}{\textwidth} 
\centerline{\includegraphics[width=0.99\temp]{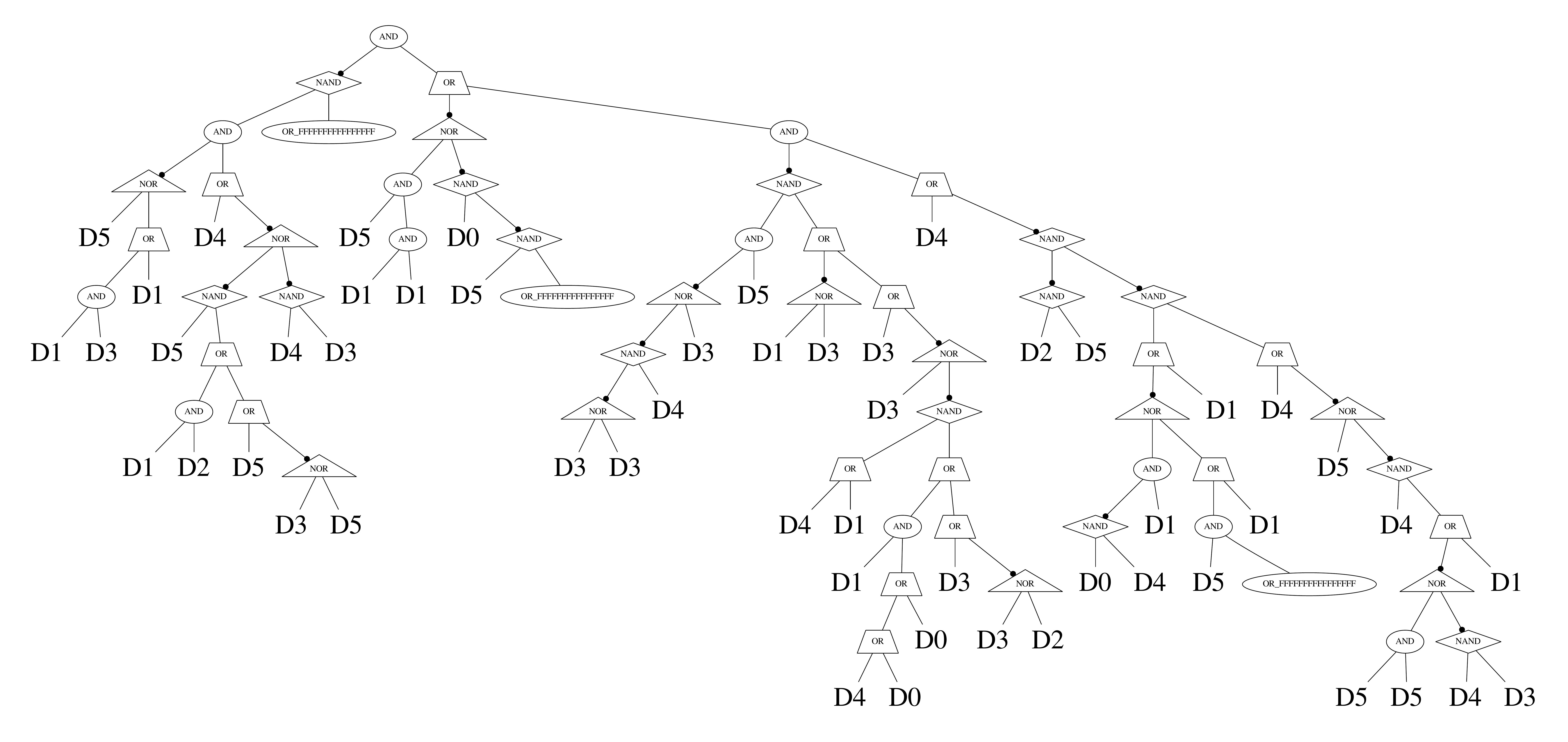}} 
\vspace*{-2ex}
\centerline{\includegraphics[width=0.99\temp]{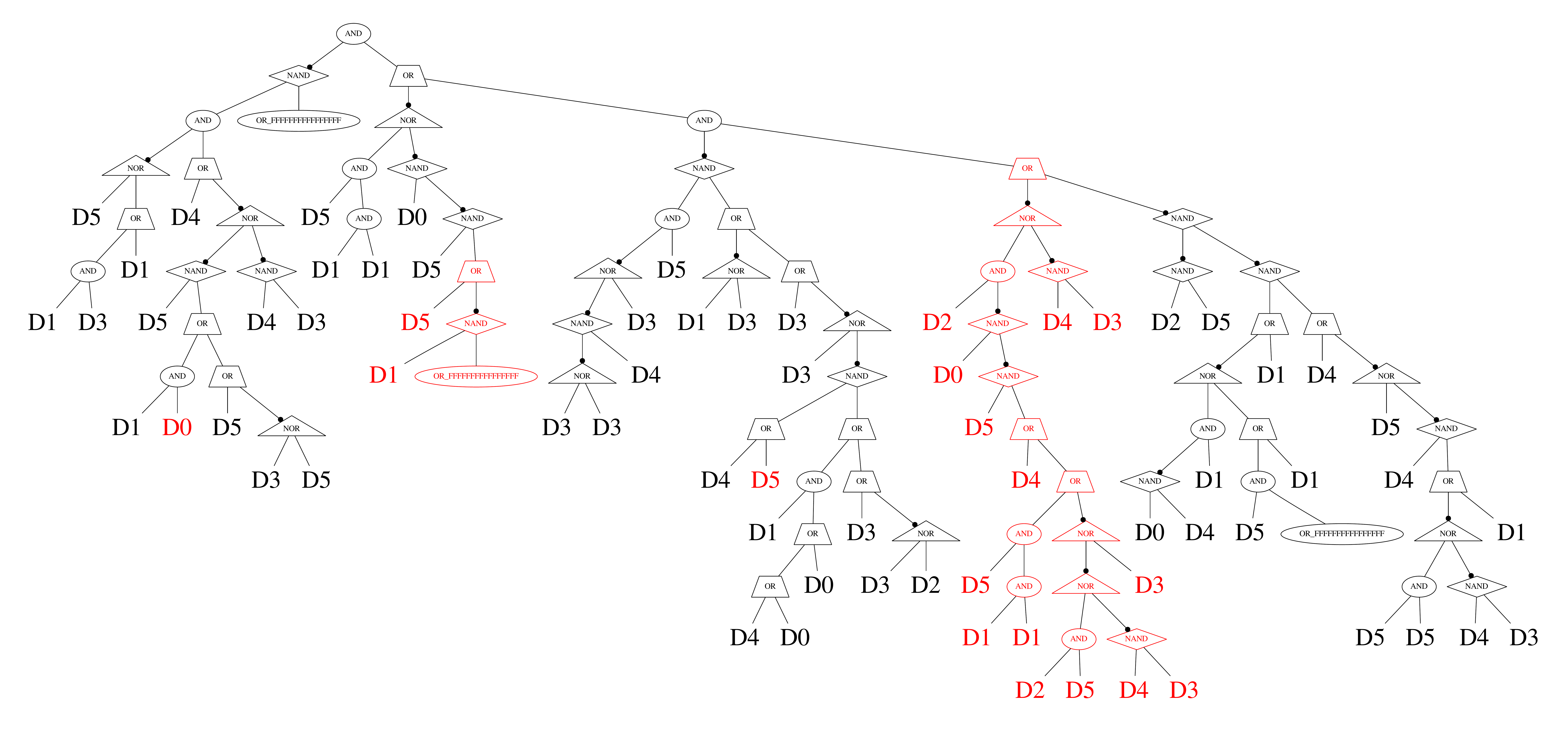}} 
\vspace*{-2ex}
\caption{\label{fig:bmux6_100_id_200000}
Typical 6-mux effective code with max fitness~(64).
Top: generation 400.
The whole program is 15\,495 instructions but only
111 are effective.
Bottom: generation 500.
Only the
141 effective out of 16\,831 instructions shown.
Notice similarity.
See also Figure~\protect\ref{fig:bmux6_100_250000}
which shows all high fitness trees in  generation 500.
}
\end{figure*}

\begin{figure*}
\begin{tabular}{@{}p{\halfwidth}@{\hspace{\columnsep}}p{\halfwidth}@{}}
\includegraphics[width=\halfwidth]{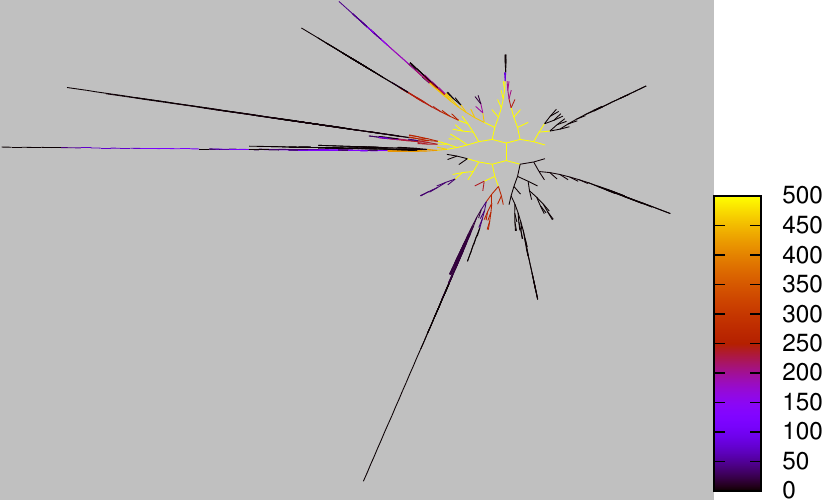} 
&
\includegraphics[width=\halfwidth]{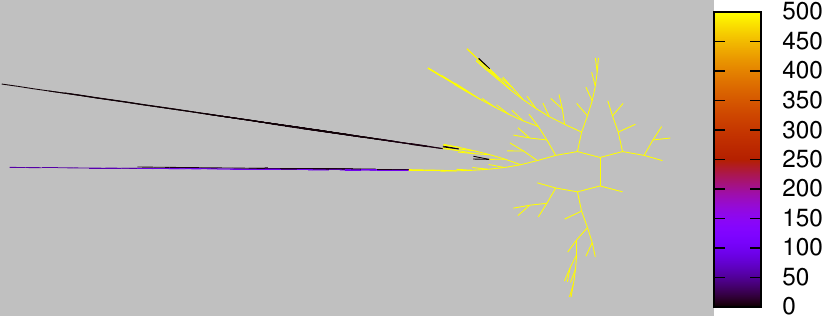} 
\\
\includegraphics[width=\halfwidth]{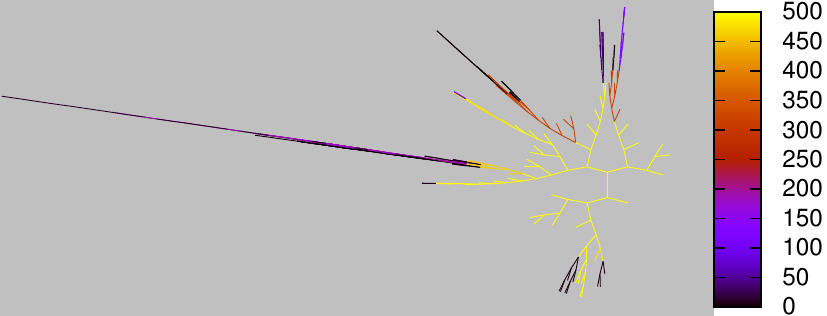} 
&
\includegraphics[width=\halfwidth]{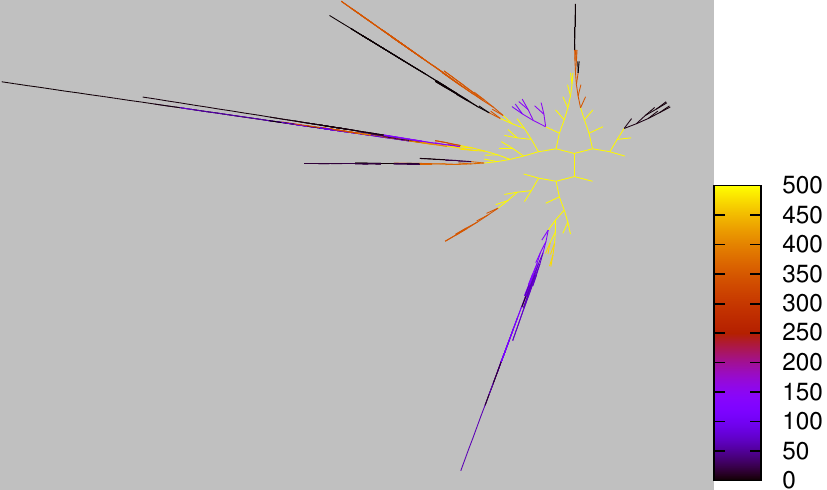} 
\end{tabular}
\vspace*{-2ex}
\caption{
\label{fig:bmux6_100_250000}
\label{fig:bmux6_100_500000}
\label{fig:bmux6_100_1000000}
\label{fig:bmux6_100_2000000}
Effective code in a population of 500 binary trees
after 500, 1000, 2000 and 4000 generations.
(Each panel showing the 496-500 trees 
with max fitness.) 
Note the similarity of the effective code even though separated by
thousands of generations.
By generation 4000
although the trees range in size from 843 to 328\,723
the effective code is limited to the
721
nodes shown 
(0.2\% of the total).
Almost all the population have 82 effective nodes in common
(yellow).
Darker colours indicate effective code 
which only occurs in $\le$148~(blue) or $\le$22~(black) trees.
Code which does not effect any program's output
is not plotted.
}
\end{figure*}

\noindent
Figure~\ref{fig:bmux6_100_3L2_size} 
(page~\pageref{fig:bmux6_100_3L2_size})
has already shown that
the size of effective code is fairly stable.
Figure~\ref{fig:bmux6_100_id_200000}
shows the effective code in two typical trees separated by 100
generations.
(I.e.\ at generation 400 and 500.)
Notice even after 100 generations the effective part of the evolved
trees is little changed.
Indeed if we look at the effective code in every tree 
in the population at generation 500
(Figure~\ref{fig:bmux6_100_250000}
page~\pageref{fig:bmux6_100_250000})
we see they are also very similar.
Typically 
the effective part of the code lies in a few hundred nodes
around the root node (yellow)
which is protected against crossover by evolved constants.
The constants head large sacrificial subtrees of ineffective code.
Figure~%
\ref{fig:bmux6_100_250000} 
shows the effective code is conserved 
over many generations.
We see this in all pop=500 runs
(except 102, gets stuck at fitness=62).
However the details of the evolved effective code differ
from run to run.

\section{A Limit to Bloat}
\label{sec:endbloat}

\noindent
Can bloat continue forever?
Even in these extended runs fitness selection is needed to sustain the
tree size \cite{Langdon:1997:bloatWSC2},
in that if we turn off selection, the number of nodes in the
population executes an apparently random walk.
However, a population of very small trees,
represents an absorbing state
from which it may take crossover a long time to escape.
Indeed 
a population of all leafs can be constructed by crossover but it
cannot escape it.
The presence of an absorbing state converts the random walk into a
gambler's ruin.

Once trees become so big that the population contains no
low fitness individuals, tree size executes a gamblers ruin
towards zero.
Although the step size increases with tree size, it
appears highly unlikely for the population to migrate to tiny trees
which crossover cannot escape,
without transitioning through a regime where small trees 
have low fitness
and hence fitness selection will kick back in to grow the
trees again.

We speculate that in a finite population
it will become possible that the bloated trees become so large that in
any generation the expected number of times
crossover disrupts the high fitness core
near the root node falls well below one per generation
Thus removing the driving force which has been growing the trees.
Hence their may be a balancing point with the gambler's ruin
near 
$$\mbox{tree size} \approx {\rm popsize} \times \mbox{core codesize}$$
In the above experiments
$500 \times 497 = 248\,500$.

\begin{figure}
\centerline{\includegraphics[scale=0.175]{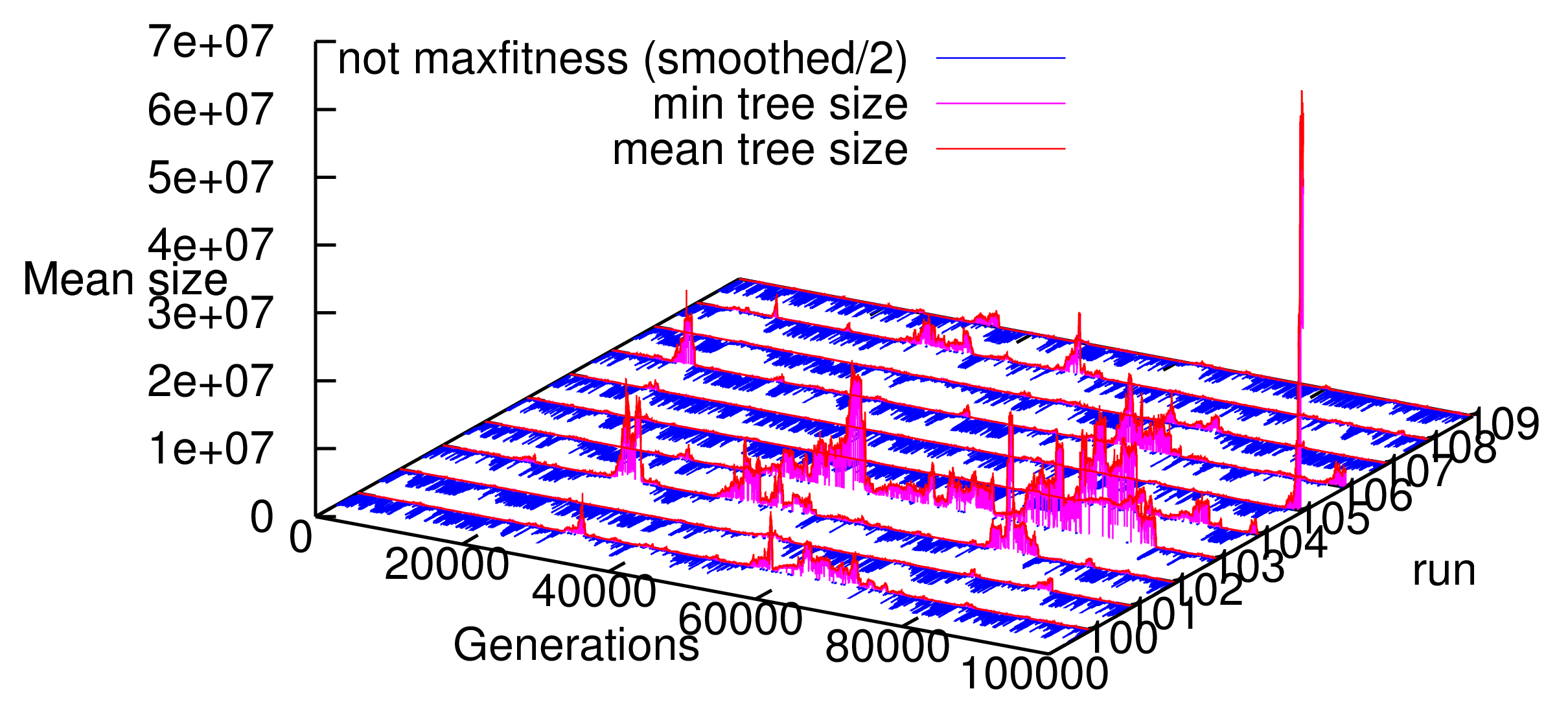}} 
\vspace*{-2ex}
\caption{
\label{fig:bmux6_p50_size}
Extended runs pop=50.
Blue horizontal 
curve shows number of GP individuals 
with 
fitness $< 64$
(rescaled by 0.5 
and
smoothed). 
In populations with huge trees,
even the smallest tree is big
and 
in many gens 
whole pop 
has same fitness.
Without a fitness differential, tree size may rise or fall.
}
\end{figure}

Figure~\ref{fig:bmux6_p50_size} 
(page~\pageref{fig:bmux6_p50_size})
shows ten extended runs with a
reduced population size and no size limits.
The smaller population size means GP is usually no longer able to solve the
problem nevertheless as expected
the run shows similar characteristics to the
larger population.
The size of the core code is not known but 
we would anticipate it would be no bigger than in runs 
where high fitness trees do evolve.
Thus we had anticipated an edge to bloat at about 25\,000.
All ten extended runs with the small population behaved similarly.
They all bloated 
(max tree size between 
3\,600\,000 
and 115\,000\,000) 
but at the end of each run the average tree size was between
0.01\% 
and 6\% of the maximum tree size.
(Only two runs reach max fitness~64.)
Across ten runs and over 100\,000 generations
the median mean tree size in the population was 
42\,507
and the median smallest was
10\,513.

\section{Conclusions}
\label{sec:conclude}

\noindent
We have studied long term evolution
(far longer than anything reported in GP).
In our populations
after thousands, even tens of thousands, of generations
trees evolve to be extremely stable so that 
there may be tens or even hundreds of generations where everyone
has the same fitness.
This means that there is no selection
and we see
bloat become an apparently random walk,
with both increases and {\em falls} in program size.
Further we suggest, 
in finite populations,
bloat is naturally limited by
a gambler's ruin process.

The evolved (albeit narrowly defined) introns
do not explain the extreme fitness convergence seen.
Instead we have described the evolution of functions with constant
output
(zero entropy).
These shelter the root node 
and are evolved to be resilient to crossover.

The evolved 
constants form a protective ring 
around highly stable effective code centred on the root node
and head huge
sacrificial subtrees of ineffective code.
(These may contain 100\,000s of useless instructions.)
This ineffective code is primarily responsible for the low number of
low fitness individuals found in highly evolved populations.

Even after evolving for thousands of generations,
in small populations,
we continue to see the impact of fitness selection on
the distribution of tree sizes.
And, although the distribution of tree sizes versus their depths
is close to that of random trees,
the distribution of tree sizes does not approach the limiting
distribution we predicted assuming no fitness~\cite{poli:2007:eurogp}.

GP code 
available via anonymous 
\href{ftp://ftp.cs.ucl.ac.uk/genetic/gp-code/GPbmux6.tar.gz}
{FTP}
and 
\href{http://www.cs.ucl.ac.uk/staff/W.Langdon/ftp/gp-code/GPbmux6.tar.gz}
{\tt http://www.cs.ucl.ac.uk/staff/W.Langdon/ftp/}
\href{http://www.cs.ucl.ac.uk/staff/W.Langdon/ftp/gp-code/GPbmux6.tar.gz}
{\tt gp-code/GPbmux6.tar.gz}

\bibliographystyle{splncs}

\bibliography{/tmp/references,/tmp/gp-bibliography}

\end{document}